\def\year{2021}\relax
\documentclass[letterpaper]{article} 
\usepackage{aaai21}  
\usepackage{times}  
\usepackage{helvet} 
\usepackage{courier}  
\usepackage[hyphens]{url}  
\usepackage{graphicx} 
\usepackage{natbib}  
\usepackage{caption} 
\usepackage{latexsym}
\usepackage{subfigure}
\usepackage{amsmath}
\usepackage{amsfonts}
\usepackage{amssymb}
\usepackage{algorithm}
\usepackage{algorithmic}
\usepackage{multirow}
\usepackage{lineno}  %
\title{KEML: A Knowledge-Enriched Meta-Learning Framework for Lexical Relation Classification}
\author {Chengyu Wang,\textsuperscript{\rm 1} Minghui Qiu,\textsuperscript{\rm 1} Jun Huang,\textsuperscript{\rm 1} Xiaofeng He\textsuperscript{\rm 2}\\}
\affiliations {
    \textsuperscript{\rm 1} Alibaba Group \\
    \textsuperscript{\rm 2} School of Computer Science and Technology, East China Normal University \\
    \{chengyu.wcy, minghui.qmh, huangjun.hj\}@alibaba-inc.com, hexf@cs.ecnu.edu.cn
}
\begin{document}

\maketitle

\begin{abstract}
Lexical relations describe how concepts are semantically related, in the form of relation triples. The accurate prediction of lexical relations between concepts is challenging, due to the sparsity of patterns indicating the existence of such relations.
We propose the Knowledge-Enriched Meta-Learning (KEML) framework to address lexical relation classification. In KEML, the LKB-BERT (Lexical Knowledge Base-BERT) model is first presented to learn concept representations from text corpora, with rich lexical knowledge injected by distant supervision. A probabilistic distribution of auxiliary tasks is defined to increase the model's ability to recognize different types of lexical relations. We further propose a neural classifier integrated with special relation recognition cells, in order to combine meta-learning over the auxiliary task distribution and supervised learning for LRC.
Experiments over multiple datasets show KEML outperforms state-of-the-art methods.
\end{abstract}

\section{Introduction}

As an important type of linguistic resources, lexical relations describe semantic associations between concepts, which are organized as backbones in lexicons, semantic networks and knowledge bases. The explicit usage of such resources has benefited a variety of NLP tasks, including relation extraction~\cite{DBLP:conf/kdd/ShenWLZRVS018}, question answering~\cite{DBLP:conf/aaai/YangZWYW17} and machine translation~\cite{DBLP:conf/emnlp/ThompsonKZKDK19}.

To accumulate such knowledge, Lexical Relation Classification (LRC) is a basic NLP task to classify concepts into a finite set of lexical relations. Pattern-based and distributional methods are two major types of LRC models~\cite{DBLP:conf/cogalex/ShwartzD16,DBLP:conf/emnlp/WangHZ17}. 
Compared to the classification of \emph{factual relations} for knowledge graph population~\cite{DBLP:conf/emnlp/LiuRZZGJH17,DBLP:conf/cikm/WuH19a}, the accurate classification of lexical relations is more challenging. i) Most lexical relations represent the \emph{common sense of human knowledge}, not frequently expressed in texts  explicitly\footnote{For example,  ``(car, meronymy, steering wheel)'' can be paraphrased as ``steering wheels are part of cars''. However, 
the expression is usually omitted in texts, since it is common sense to humans.}. 
Apart from Hearst patterns~\cite{DBLP:conf/coling/Hearst92} for hypernymy (``\emph{is-a}'') extraction, textual patterns that indicate the existence of other types of lexical relations remain few, leading to the ``\emph{pattern sparsity}'' problem~\cite{DBLP:conf/cogalex/ShwartzD16,DBLP:conf/naacl/WashioK18}. ii) Distributional models assume concepts with similar contexts have similar embeddings~\cite{DBLP:journals/corr/abs-1301-3781}.
Representations of a concept pair learned by traditional word embedding models are not sufficient to distinguish different types of lexical relations~\cite{DBLP:conf/naacl/GlavasV18,DBLP:conf/emnlp/PontiVGRK19}.
iii) Many LRC datasets are \emph{highly imbalanced} w.r.t. training instances of different lexical relations, and may contain \emph{randomly paired} concepts.
It is difficult for models to distinguish whether a concept pair has a particular type of lexical relation, or has \emph{very weak or no semantic relatedness}. 

In this paper, we present the \emph{Knowledge-Enriched Meta-Learning} (KEML) framework to address these challenges of LRC. KEML consists of three modules: \emph{Knowledge Encoder}, \emph{Auxiliary Task Generator} and \emph{Relation Learner}.
In \emph{Knowledge Encoder}, we propose the LKB-BERT (Lexical Knowledge Base-BERT) model to learn \emph{relation-sensitive concept representations}. LKB-BERT is built upon BERT~\cite{DBLP:conf/naacl/DevlinCLT19} and trained via \emph{new distantly supervised learning tasks} over lexical knowledge bases, encoding both contextual representations and relational lexical knowledge. In \emph{Auxiliary Task Generator}, we treat recognizing single type of lexical relations as \emph{auxiliary tasks}. Based on meta-learning~\cite{DBLP:conf/icml/FinnAL17}, a probabilistic distribution of auxiliary tasks is properly defined for the model to optimize, which addresses the imbalanced property and the existence of random relations in LRC datasets. In \emph{Relation Learner}, we design a feed-forward neural  network for LRC, of which the training process combines both gradient-based meta-learning and supervised learning. Especially, a special relation recognition cell is designed and integrated into the network for the purpose.

This paper makes the following contributions:
\begin{itemize}
\item We present LKB-BERT to learn relation-sensitive contextual representations of concepts by leveraging rich relational knowledge in knowledge bases.

\item A meta-learning process combined with an auxiliary task distribution is defined to improve the ability of distributional LRC models to recognize lexical relations.

\item We design a feed-forward neural network combined with special relation recognition cells to accommodate both meta-learning and supervised learning for LRC.

\item In the experiments, we evaluate the KEML model over multiple public benchmark datasets. Experimental results show that KEML outperforms state-of-the-art methods.
\end{itemize}


\section{Related Work}
\label{sec:related}

In this section, we overview the related work on LRC, pre-trained language models and meta-learning.

\noindent\textbf{Lexical Relation Classification.}
As summarized in~\citet{DBLP:conf/cogalex/ShwartzD16}, LRC models are categorized into two major types: \emph{pattern-based} and \emph{distributional}. Pattern-based approaches extract patterns w.r.t. a concept pair from texts as features to predict its lexical relation. For hypernymy relations, Hearst patterns~\cite{DBLP:conf/coling/Hearst92} are most influential, often used for the construction of large-scale taxonomies~\cite{DBLP:conf/sigmod/WuLWZ12}. To learn patterns representations,~\citet{DBLP:conf/acl/ShwartzGD16} exploit LSTM-based RNNs to encode dependency paths of patterns.~\citet{DBLP:conf/acl/RollerKN18,DBLP:conf/acl/LeRPKN19} calculate Hearst pattern-based statistics from texts to predict the degrees of hypernymy between concepts. For other types of lexical relations, LexNET~\cite{DBLP:conf/cogalex/ShwartzD16} extends the network architecture~\cite{DBLP:conf/acl/ShwartzGD16} for multi-way classification of lexical relations.~\citet{DBLP:conf/acl/NguyenWV16,DBLP:conf/eacl/NguyenWV17} design path-based neural networks to distinguish antonymy and synonymy.
However, these methods may suffer from the lack of patterns and concept pair occurrence in texts~\cite{DBLP:conf/naacl/WashioK18}.

With the rapid development of deep learning, distributional models attract more interest.
While traditional methods directly leverage the two concepts' embeddings as features for the classifier~\cite{DBLP:conf/coling/WeedsCRWK14,DBLP:conf/acl/VylomovaRCB16,DBLP:conf/emnlp/RollerE16}, they may suffer from the ``\emph{lexical memorization}'' problem~\cite{DBLP:conf/naacl/LevyRBD15}. 
Recently, more complicated neural networks are proposed.~\citet{DBLP:conf/cogalex/AttiaMSKS16} formulate LRC as a multi-task learning task and propose a convolutional neural network for LRC.
~\citet{DBLP:journals/tacl/MrksicVSLRGKY17} propose the Attract-Repe model to learn the semantic specialization of word embeddings.
~\citet{DBLP:conf/naacl/GlavasV18} introduce the Specialization Tensor Model, which learns multiple relation-sensitive specializations of concept embeddings.
SphereRE~\cite{DBLP:conf/acl/WangHZ19} encodes concept pairs in the hyperspherical embedding space and achieves state-of-the-art results.
A few works learn word-pair representations for other NLP tasks~\cite{DBLP:conf/emnlp/WashioK18,DBLP:conf/naacl/JoshiCLWZ19,DBLP:conf/acl/Camacho-Collados19}.
KEML is also distributional, improving LRC by training meta-learners over language models.

\noindent\textbf{Pre-trained Language Models.}
Pre-trained language models have gained attention from the NLP community~\cite{DBLP:journals/corr/abs-2003-08271}.
ELMo~\cite{DBLP:conf/naacl/PetersNIGCLZ18} learns context-sensitive embeddings for each token form both left-to-right and right-to-left. BERT~\cite{DBLP:conf/naacl/DevlinCLT19} employs layers of transformer encoders to learn language representations. Follow-up works include Transformer-XL~\cite{DBLP:conf/acl/DaiYYCLS19}, XLNet~\cite{DBLP:conf/nips/YangDYCSL19}, ALBERT~\cite{DBLP:journals/corr/abs-1909-11942} and many more.
Yet another direction is to fuse additional knowledge sources into BERT-like models. 
ERNIE~\cite{DBLP:conf/acl/ZhangHLJSL19} incorporates the rich semantics of entities in the model. KG-BERT~\cite{DBLP:journals/corr/abs-1909-03193} and K-BERT~\cite{DBLP:journals/corr/abs-1909-07606} employ relation prediction objectives in knowledge graphs as additional learning tasks.~\citet{DBLP:conf/iclr/XiongDWS20} propose to pre-train neural language models with online encyclopedias. In our work, we leverage the conceptual facts in lexical knowledge bases to improve the representation learning for LRC.

\noindent\textbf{Meta-learning.}
Meta-learning is a learning paradigm to train models that can adapt to a variety of different tasks with little training data~\cite{DBLP:journals/corr/abs-1810-03548}, mostly applied to few-shot learning (typically by N-way K-shot). In NLP, meta-learning algorithms have not been extensively employed, mostly due to the large numbers of training examples required to train the model for different NLP tasks. Existing models mostly focus on training meta-learners for single applications, such as link prediction~\cite{DBLP:conf/emnlp/ChenZZCC19}, dialog systems~\cite{DBLP:conf/acl/MadottoLWF19} and semantic parsing~\cite{DBLP:conf/acl/GuoTDZY19}.~\citet{DBLP:conf/emnlp/DouYA19} leverage meta-learning for various low-resource natural language understanding tasks. 

To our knowledge, KEML is one of the early attempts to improve LRC by meta-learning~\cite{DBLP:conf/icml/FinnAL17}. Different from the traditional N-way K-shot setting, our work employs the meta-learning framework as an intermediate step to enhance the capacity of the relation classifier. Since the mechanism of meta-learning is not our major research focus, we do not further elaborate.

\begin{figure*}
\centering
\includegraphics[width=.75\textwidth]{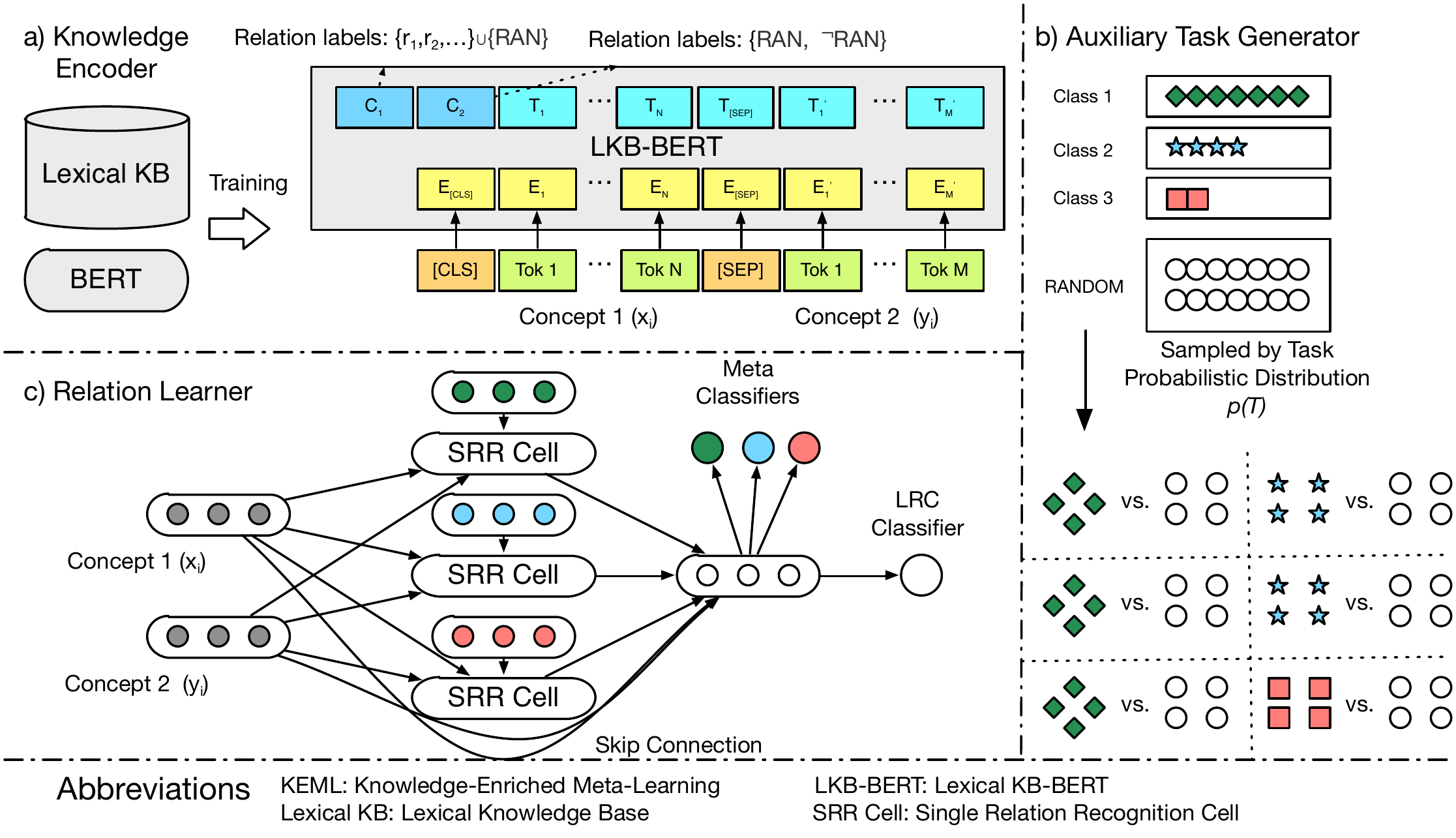}
\caption{The high-level framework of KEML (best viewed in color).}
\label{fig:example}
\end{figure*}

\section{The KEML Framework}
\label{sec:framework}

In this section, we formally describe the KEML framework. A brief technical flow is presented, followed by its details.

\subsection{A Brief Overview of KEML}

We first overview the LRC task briefly. Denote $(x_i,y_i)$ as an arbitrary concept pair. The goal of LRC is to learn a classifier $f$ to predict the lexical relation $r_i\in\mathcal{R}$ between $x_i$ and $y_i$, based on a labeled, training set $\mathcal{D}=\{(x_i,y_i,r_i)\}$. Here, $\mathcal{R}$ is the collection of all pre-defined lexical relation types (e.g., hypernymy, synonymy), (possibly) including a special relation type $\mathbf{RAN}$ (``\emph{random}''), depending on different task settings. It means that $x_i$ and $y_i$ are randomly paired, without clear association with any lexical relations.

The framework of KEML is illustrated in Figure~\ref{fig:example}, with three modules introduced below:

\noindent\textbf{Knowledge Encoder.}
Representation learning for LRC is significantly different from learning traditional word embeddings. This is because some lexical concepts are naturally \emph{Multiword Expressions} (e.g., \emph{card game}, \emph{orange juice}), in which the entire sequences of tokens should be encoded in the embedding space~\cite{DBLP:conf/acl/CordeiroRIV16}. Additionally, these models are insufficient to capture the lexical relations between concepts, due to the pattern sparsity issue~\cite{DBLP:conf/naacl/WashioK18}. Hence, the semantics of concepts should be encoded from a larger corpus and rich language resources.
Inspired by BERT~\cite{DBLP:conf/naacl/DevlinCLT19}, and its extensions,
we propose the LKB-BERT (Lexical Knowledge Base-BERT) model to encode the semantics of concepts from massive text corpora and lexical knowledge bases. After the training of LKB-BERT, each concept $x_i$ or $y_i$ receives the embeddings of the last transformer encoding layer of LKB-BERT as its representation. Denote the embeddings of $x_i$ (or $y_i$) as $\vec{x}_i$ (or $\vec{y}_i$), with the dimension as $d$.


\textbf{Auxiliary Task Generator.}
As discussed in related work, training embedding based classifiers directly may produce sub-optimal results due to i) the~\emph{lexical memorization} problem~\cite{DBLP:conf/naacl/LevyRBD15} and ii) the \emph{highly imbalanced} nature of LRC training sets and the existence of $\mathbf{RAN}$ relations. 
Inspired by the design philosophy of meta-learning~\cite{DBLP:conf/icml/FinnAL17,DBLP:conf/nips/FinnXL18} and its NLP applications~\cite{DBLP:conf/emnlp/ChenZZCC19,DBLP:conf/acl/MadottoLWF19}, before we learn the LRC model, we design a series of auxiliary tasks for the model to solve.
Each task aims at distinguishing between concept pairs that have a particular relation $r\in\mathcal{R}$ and randomly paired concepts. 
By solving these tasks, the network can gradually learn how to identify each type of lexical relation $r$ before the final relation classifier is trained. The training procedure is discussed next.

\textbf{Relation Learner.}
We design a two-stage algorithm to train a neural network based relation classifier $f$: i) meta-learning and ii) supervised learning.
In the meta-learning stage,  the adapted model parameters are iteratively learned by solving auxiliary tasks. Hence, the neural network learns how to recognize specific lexical relations, with the guidance of the lexical knowledge base. 
In the supervised learning stage, we fine-tune meta-learned parameters to obtain the multi-way classification model for LRC over $\mathcal{D}$.

\subsection{Knowledge Encoder}

The LKB-BERT model has the similar network architecture to that of BERT~\cite{DBLP:conf/naacl/DevlinCLT19}. In original BERT, the inputs are arbitrary spans of token sequences. To encode the semantics of concept pairs, we combine a concept pair $(x_i,y_i)$ to form a sequence of tokens, separated by a special token ``[SEP]'' as the input for LKB-BERT (see Figure~\ref{fig:example}). We first initialize all the model parameters of transformer encoders to be the same as BERT's pre-training results (apart from the output layer). Different from any standard fine-tuning tasks in BERT~\cite{DBLP:conf/naacl/DevlinCLT19}, to address the $\mathbf{RAN}$ problem, LKB-BERT has two output classifiers.
The first one classifies a concept pair $(x_i,y_i)$ into the lexical relation collection $\mathcal{R}$ (including the special relation type $\mathbf{RAN}$). 
Let $\mathcal{KB}$ be the collection of labeled concept pairs in lexical knowledge bases. For each concept pair with its label $(x_i,y_i,r_i)\in\mathcal{KB}$, we compute $\tau_r(x_i,y_i)$ as the predicted score w.r.t. the lexical relation $r$ by LKB-BERT's transformer encoders (we have $\forall r\in\mathcal{R}, \tau_{r}(x_i,y_i)\in[0,1]$ and $\sum_{r\in\mathcal{R}}\tau_{r}(x_i,y_i)=1$).
The first loss, i.e., the multi-way relation classification loss $\mathcal{L}_{KB}^{(1)}$ is defined as follows:\footnote{For simplicity, we omit all the regularization terms of objective functions throughout this paper.}
\begin{equation*}
\mathcal{L}_{KB}^{(1)}=-\sum_{(x_i,y_i,r_i)\in\mathcal{KB}}\sum_{r\in\mathcal{R}}(\mathbf{1}(r_i=r)\cdot\log\tau_{r}(x_i,y_i))
\end{equation*}
where $\mathbf{1}(\cdot)$ is the indicator function that returns 1 if the input expression is true; and 0 otherwise.
 
To improve LKB-BERT's ability to recognize concept pairs without any lexical relations, we add a binary cross-entropy loss for LKB-BERT to optimize. We only need LKB-BERT to learn whether a concept pair $(x_i,y_i)$ are randomly paired. Let $\mathbf{^\neg RAN}$ be any non-random lexical relation type in $\mathcal{R}$. The complete objective of LKB-BERT is: $\mathcal{L}_{KB}=\mathcal{L}_{KB}^{(1)}+\mathcal{L}_{KB}^{(2)}$, with $\mathcal{L}_{KB}^{(2)}$ to be:
\begin{equation*}
\begin{split}
\mathcal{L}_{KB}^{(2)}=&-\sum_{(x_i,y_i,r_i)\in\mathcal{KB}}(\mathbf{1}(r_i=\mathbf{RAN})\cdot\log\tau_{\mathbf{RAN}}(x_i,y_i)\\ 
&+\mathbf{1}(r_i=\mathbf{^\neg RAN})\cdot\log\tau_{\mathbf{^\neg RAN}}(x_i,y_i) )
\end{split}
\end{equation*}
In KEML, we regard lexical relations sampled from WordNet~\cite{DBLP:journals/cacm/Miller95} and the training set that we use to train the final relation classifier as the sources of $\mathcal{KB}$.
Readers can also refer to the C$_1$ and C$_2$ units of LKB-BERT in Figure~\ref{fig:example}.

\subsection{Auxiliary Task Generator}

Although LKB-BERT is capable of learning deep concept representations, using such features for classifier training is insufficient. The reasons are twofold. i) Direct classification can suffer from ``lexical memorization''~\cite{DBLP:conf/naacl/LevyRBD15}, meaning that the relation classifier $f$ only learns the individual characteristics of each one of the two concepts alone.
ii) The LRC datasets are highly imbalanced. For example, in the widely used dataset EVALution~\cite{DBLP:conf/acl-ldl/SantusYLH15}, the numbers of training instances w.r.t. several lexical relation types are very few. Hence, the learning bias of the classifier trained by naive approaches is almost unavoidable.

The small number of training instances of certain lexical relations motivates us to employ few-shot meta-learning to address this issue~\cite{DBLP:conf/icml/FinnAL17}, instead of multi-task learning of all the lexical relations.
In KEML, we propose a series of auxiliary tasks $\mathcal{T}$, where each task (named \emph{Single Relation Recognition}) $\mathcal{T}_{r}\in\mathcal{T}$ corresponds to a specific type of lexical relation $r\in\mathcal{R}$ (excluding $\mathbf{RAN}$). The goal is to distinguish concept pairs with the lexical relation type $r$ and randomly paired concepts. Let $\mathcal{S}_{r}$ and $\mathcal{S}_{\mathbf{RAN}}$ be the collection of concept pairs with lexical relations as $r$ and $\mathbf{RAN}$, respectively, randomly sampled from the training set $\mathcal{D}$. The goal of learning auxiliary task $\mathcal{T}_{r}$ is to minimize the following loss function $\mathcal{L}({\mathcal{T}_{r}})$:
\begin{equation*}
\begin{split}
&\mathcal{L}({\mathcal{T}_{r}})= -\sum_{(x_i,y_i,r_i)\in\mathcal{S}_r\cup\mathcal{S}_{\mathbf{RAN}}}(\mathbf{1}(r_i=r)\\
&\cdot\log q_{r}(x_i,y_i)+\mathbf{1}(r_i=\mathbf{RAN})\cdot\log q_{\mathbf{RAN}}(x_i,y_i) )
\end{split}
\end{equation*}
where $q_r(x_i,y_i)$ is the predicted probability of the concept pair $(x_i,y_i)$ having the lexical relation $r$. By solving the tasks $\mathcal{T}$, the model gradually learns how to recognize all these lexical relations, and to identify whether there exists any lexical relation between the two concepts at all.

A remaining problem is the design of the probabilistic distribution of auxiliary tasks $p(\mathcal{T})$. 
We need to consider two issues. i) The semantics of all types of lexical relations should be fully learned. ii) Assume the batch sizes for all tasks are the same, i.e., $\forall r_p,r_q\in\mathcal{R}\setminus\{\mathbf{RAN}\}, \vert\mathcal{S}_{r_p}\vert=\vert\mathcal{S}_{r_q}\vert$. Tasks related to lexical relations with more training instances should be learned more frequently by the meta-learner. Let $\mathcal{D}_{r}$ be the subset of the training set $\mathcal{D}$ with the lexical relation as $r$.
$\forall r_p,r_q\in\mathcal{R}\setminus\{\mathbf{RAN}\}$, if $\vert\mathcal{D}_{r_p}\vert>\vert\mathcal{D}_{r_q}\vert$, we have the sampling probability $p(\mathcal{T}_{r_p})>p(\mathcal{T}_{r_q})$.
Hence, we define $p(\mathcal{T}_{r_p})$ empirically as follows:
\begin{equation*}
p(\mathcal{T}_{r})=\frac{\ln\vert \mathcal{D}_{r}\vert +\gamma}{\sum_{r^{'}\in\mathcal{R}\setminus\{\mathbf{RAN}\}}(\ln\vert \mathcal{D}_{r^{'}}\vert +\gamma)}
\end{equation*}
where $\gamma>0$ is the smoothing factor. This setting over-samples lexical relations with few training instances, alleviating the imbalanced class issue.
Overall, the expectation of all the losses of auxiliary tasks (represented as $\mathcal{L}(\mathcal{T})$) is: $\mathbb{E}(\mathcal{L}(\mathcal{T}))=\sum_{r^*\in\mathcal{R}\setminus\{\mathbf{RAN}\}}p(\mathcal{T}_{r})\cdot \mathcal{L}(\mathcal{T}_{r})$, which is the real learning objective that these auxiliary tasks aim to optimize. 

\subsection{Relation Learner}
We now introduce the learning algorithm and the network structure for LRC. 
Assume the classifier $f$ is parameterized by $\theta$, with learning and meta-learning rates as $\alpha$ and $\epsilon$.
Relation Learner has two stages: i) meta-learning and ii) supervised learning. For each iteration in meta-learning, we sample $N$ auxiliary tasks from $p(\mathcal{T})$. For each auxiliary task $\mathcal{T}_{r}$, we learn adapted parameters based on two sampled subsets: $\mathcal{S}_r$ and $\mathcal{S}_{\mathbf{RAN}}$, to make the model to recognize one specific relation type. After that, the adapted parameters on each task $\mathcal{T}_{r}$ are averaged and updated to $\theta$. We simplify the meta-update step by only taking first-order derivatives~\cite{DBLP:journals/corr/abs-1803-02999} to avoid the time-consuming second-order derivative computation. 
For supervised learning, we fine-tune the parameters $\theta$ of the classifier $f$ to obtain the multi-way LRC model over the entire training set $\mathcal{D}$.
The algorithmic description is shown in Algorithm~\ref{alg:meta}.

\begin{algorithm}[h]
\caption{Meta-Learning Algorithm for LRC}
\label{alg:meta}
\begin{small}
\begin{algorithmic}[1]
\STATE Initialize model parameters $\theta$;
\WHILE{not converge}
\STATE Sample $N$ auxiliary tasks $\mathcal{T}_{r_1}, \mathcal{T}_{r_2},\cdots, \mathcal{T}_{r_N}$ from the task distribution $p(\mathcal{T})$;
\FOR {each auxiliary task $\mathcal{T}_{r}$}
\STATE Sample a batch (positive samples $\mathcal{S}_r$ and negative samples $\mathcal{S}_{\mathbf{RAN}}$) from the training set $\mathcal{D}$;
\STATE Update adapted parameters: $\theta_r\leftarrow\theta-\alpha\nabla\mathcal{L}({\mathcal{T}_{r}})$ based on $\mathcal{S}_r$ and $\mathcal{S}_{\mathbf{RAN}}$;
\ENDFOR
\STATE Update meta-parameters: $\theta\leftarrow\theta-\epsilon\nabla\sum_{\mathcal{T}_{r}}\mathcal{L}({\mathcal{T}_{r}})$;
\ENDWHILE
\STATE Fine-tune $\theta$ over $\mathcal{D}$ by standard supervised learning LRC;
\end{algorithmic}
\end{small}
\end{algorithm}

\begin{figure}[ht]
\centering
\includegraphics[width=.4\textwidth]{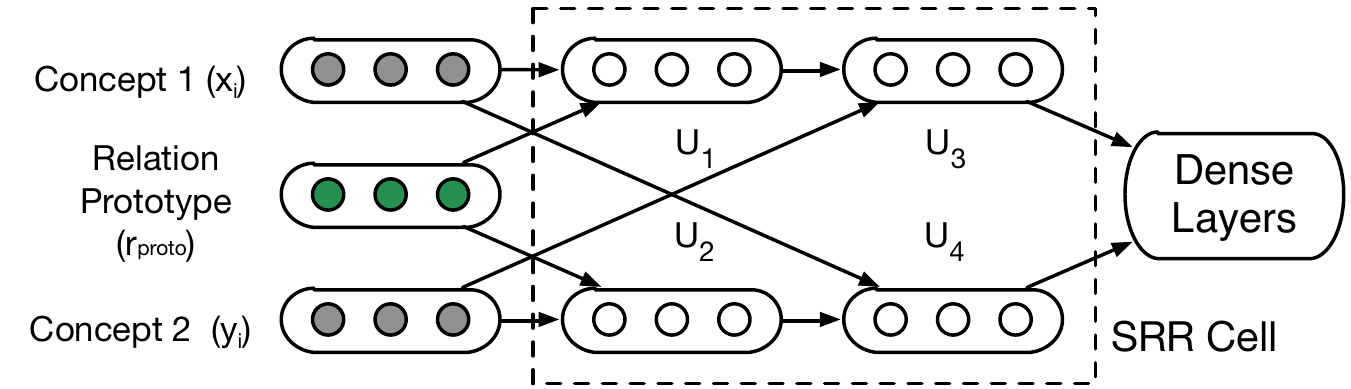}
\caption{Structure of SRR Cell (we only show one cell, with some other parts of the network omitted).}
\label{fig:cell}
\end{figure}

Finally, we describe the neural network structure for LRC.
In this network, the \emph{Single Relation Recognition Cell} (SRR Cell) is designed for learning auxiliary tasks and enabling knowledge injection, with the structure illustrated in Figure~\ref{fig:cell}.
For each lexical relation $r\in\mathcal{R}\setminus\{\mathbf{RAN}\}$, we extract the \emph{relation prototype} $\vec{r}_{proto}$ from the lexical knowledge base $\mathcal{KB}$ by averaging all the embedding offsets (i.e., $\vec{x}_i-\vec{y}_i$) of concept pairs $(x_i,y_i)$ with relation $r$:
\begin{equation*}
\vec{r}_{proto}=\frac{\sum_{(x_i,y_i,r_i)\in\mathcal{KB}} (\mathbf{1}(r_i=r)\cdot(\vec{x}_i-\vec{y}_i))}{\sum_{(x_i,y_i,r_i)\in\mathcal{KB}}\mathbf{1}(r_i=r)}
\end{equation*}
We use $\vec{x}_i-\vec{y}_i$ as features because the \emph{Diff} model is effective for representing semantic relations~\cite{DBLP:conf/acl/FuGQCWL14,DBLP:conf/acl/VylomovaRCB16,DBLP:conf/acl/WangHZ19}. Consider the SRR Cell structure in Figure~\ref{fig:cell}. Given the inputs $\vec{x}_i$, $\vec{y}_i$ and $\vec{r}_{proto}$, we compute the hidden states $\vec{U}_1$ and $\vec{U}_2$ by:
\begin{equation*}
\vec{U}_1=\tanh((\vec{x}_i\oplus\vec{r}_{proto})\cdot\mathbf{W}_1+\vec{b}_1)
\end{equation*}
\begin{equation*}
\vec{U}_2=\tanh((\vec{y}_i\oplus\vec{r}_{proto})\cdot\mathbf{W}_2+\vec{b}_2)
\end{equation*}
where $\mathbf{W}_1\in\mathbb{R}^{2d\times d}$, $\mathbf{W}_2\in\mathbb{R}^{2d\times d}$, $\vec{b}_1\in\mathbb{R}^{d}$ and $\vec{b}_2\in\mathbb{R}^{d}$ are the weights and biases of these hidden states. This can be interpreted as \emph{inferring the embeddings of relation objects or subjects}, given the relation prototype and subjects/objects as inputs, similar to knowledge graph completion~\cite{DBLP:journals/tkde/WangMWG17}.
Next, we compute the offsets $\vec{U}_1-\vec{y}_i$ and $\vec{U}_2-\vec{x}_i$ and two new $d$-dimensional hidden states $\vec{U}_3$ and $\vec{U}_4$, with $\mathbf{W}_3\in\mathbb{R}^{d\times d}$, $\mathbf{W}_4\in\mathbb{R}^{d\times d}$, $\vec{b}_3\in\mathbb{R}^{d}$ and $\vec{b}_4\in\mathbb{R}^{d}$ as learnable parameters. $\vec{U}_3$ and $\vec{U}_4$ are defined as:
\begin{equation*}
\vec{U}_3=\tanh((\vec{U}_1-\vec{y}_i)\cdot\mathbf{W}_3+\vec{b}_3)
\end{equation*}
\begin{equation*}
\vec{U}_4=\tanh((\vec{U}_2-\vec{x}_i)\cdot\mathbf{W}_4+\vec{b}_4)
\end{equation*}

We can see that if $x_i$ and $y_i$ actually have the lexical relation $r$, $\vec{U}_3$ and $\vec{U}_4$ should be good indicators of the existence of such relations. For example, one way to interpret $\vec{U}_1$ and $\vec{U}_3$ is that $\vec{U}_1$ tries to infer the relation object given $\vec{x}_i$ and $\vec{r}_{proto}$ as inputs, and $\vec{U}_3$ makes the judgment by comparing $\vec{U}_1$ and the true relation object embedding $\vec{y}_i$. Hence, the network learns whether $\vec{r}_{proto}$ is a good fit for the concept pair $(x_i,y_i)$. The functionalities of $\vec{U}_2$ and $\vec{U}_4$ are similar, only with directions reversed. After that, we concatenate $\vec{U}_3$ and $\vec{U}_4$ as part of the inputs for the next layer.

Re-consider the entire network structure in Figure~\ref{fig:example}. For each concept pair $(x_i,y_i)$, we compare $\vec{x}_i$ and $\vec{y}_i$ with all relation prototypes (treated as constants in the network). The results are represented by $2(\vert\mathcal{R}\vert-1)$ vectors of hidden states.
After that, a dense layer and multiple output layers are connected. During the meta-learning stage, we train $\vert\mathcal{R}\vert-1$ binary meta-classifiers to minimize $\mathcal{L}(\mathcal{T}_r)$ ($\mathcal{T}_r\in\mathcal{T}$), with meta-parameters updated. In the supervised learning stage, we discard all the output layers of meta-classifiers, and train the final LRC model. This is because the numbers of output units of meta-classifiers and the final classifier are different. The parameters of the last layer can not be re-used. We also need to note that additional \emph{skip connections} between $\vec{x}_i$, $\vec{y}_i$ and the dense layer are added, in order to improve the effect of back propagation during training~\cite{DBLP:conf/cvpr/HeZRS16}. 
 
\noindent\textbf{Discussion.} Overall speaking, KEML employs a ``divide-and-conquer'' strategy for LRC. In the meta-learning step, each SRR Cell learns the semantics of single lexical relation type, and also handles the $\mathbf{RAN}$ problem. Hence, the supervised learning process of the relation classifier can be improved with better parameter initializations, as model parameters already have some knowledge about relation recognition. Unlike traditional meta-learning~\cite{DBLP:conf/icml/FinnAL17,DBLP:conf/nips/FinnXL18}, KEML does not contain meta-testing steps (since LRC is not an N-way K-shot learning problem), but takes advantages of meta-learning as an intermediate step for the supervised training of the relation classifier afterwards.


\begin{table*}  
\begin{small}
\begin{center}
\begin{tabular}{l lll lll lll lll}  
\hline 
\multirow{2}{1cm}{\bf Method} &  \multicolumn{3}{c}{K\&H+N} & \multicolumn{3}{c}{BLESS} & \multicolumn{3}{c}{ROOT09} & \multicolumn{3}{c}{EVALution}\\ 
 & Pre & Rec & F1 & Pre & Rec & F1 & Pre & Rec & F1 & Pre & Rec & F1\\
\hline
Concat & 0.909 & 0.906 & 0.904 & 0.811 & 0.812 & 0.811 & 0.636 & 0.675 & 0.646 & 0.531 & 0.544 & 0.525\\
Diff & 0.888 & 0.886 & 0.885 & 0.801 & 0.803 & 0.802 & 0.627 & 0.655 & 0.638 & 0.521 & 0.531 & 0.528\\
NPB & 0.713 & 0.604 & 0.55 & 0.759 & 0.756 & 0.755 & 0.788 & 0.789 & 0.788 & 0.53 & 0.537 & 0.503\\
NPB+Aug & - & - & 0.897 & - & - & 0.842 & - & - & 0.778 & - & - & 0.489\\
LexNET & 0.985 & 0.986 & 0.985 & 0.894 & 0.893 & 0.893 & 0.813 & 0.814 & 0.813 & 0.601 & 0.607 & 0.6\\
LexNET+Aug & - & - & 0.970 & - & - & 0.927 & - & - & 0.806 & - & - & 0.545\\
SphereRE & 0.990 & 0.989 & 0.990 & 0.938 & 0.938 & 0.938 & 0.860 & 0.862 & 0.861 & 0.62 & 0.621 & 0.62\\
\hline
\bf LKB-BERT & 0.981 & 0.982 & 0.981 & \bf 0.939 & 0.936 & 0.937 & \bf 0.863 & \bf 0.864 & \bf 0.863 & \bf 0.638 & \bf 0.645 & \bf 0.639\\
\bf KEML-S & 0.984 & 0.983 &  0.984 &  \bf 0.942 &  \bf 0.940 & \bf 0.941 & \bf 0.877 & \bf 0.871 & \bf 0.873 & \bf 0.649 & \bf 0.651 & \bf 0.644\\
\bf KEML & \bf 0.993 & \bf 0.993 & \bf 0.993 & \bf 0.944 & \bf 0.943 & \bf 0.944 & \bf 0.878 & \bf 0.877 & \bf 0.878 & \bf 0.663 & \bf 0.660 & \bf 0.660\\
\hline 
\end{tabular}
\end{center}
\end{small} 
\caption{LRC results over four benchmark datasets in terms of Precision, Recall and F1.} \label{tab:f1} 
\end{table*}

\section{Experiments}
\label{sec:exp}

In this section, we conduct extensive experiments to evaluate KEML over multiple benchmark datasets, and compare it with state-of-the-arts to make the convincing conclusion.

\begin{table}  
\centering
\begin{scriptsize} 
\begin{tabular}{lllllll}  
\hline 
\bf Relation & K\&H+N & BLESS & ROOT09 & EVAL. & Cog.\\
\hline
Antonym & - & - & - & 1600 & 601\\
Attribute & - & 2731 & - & 1297 & -\\
Co-hyponym & 25796 & 3565 & 3200 & - & -\\
Event & - & 3824 & - & - & -\\
Holonym & - & - & - & 544 & -\\
Hypernym & 4292 & 1337 & 3190 & 1880 & 637\\
Meronym & 1043 & 2943 & - & 654 & 387\\
Random & 26378 & 12146 & 6372 & - & 5287\\
Substance & - & - & - & 317 & -\\
Meronym\\
Synonym & - & - & - & 1086 & 402\\
\hline 
\bf Total & 57509 & 26546 & 12762 & 7378 & 7314\\
\hline
\end{tabular}
\end{scriptsize} 
\caption{Dataset statistics. Relation names in all the datasets have been mapped to relation names in WordNet. ``EVAL.'' and ``Cog.'' are short for ``EVALution'' and ``CogALex''.} \label{tab:stat} 
\end{table}

\subsection{Datasets and Experimental Settings}

We employ Google's pre-trained BERT model\footnote{We use the uncased, base version of BERT. See~\url{https://github.com/google-research/bert}.} to initialize the parameters of LKB-BERT. The lexical knowledge base $\mathcal{KB}$ contains 16.7K relation triples~\footnote{To avoid data leakage, we have removed relation triples in~$\mathcal{KB}$ such that at least one word (either the relation subject or object) overlaps with relation triples in validation and testing sets. This setting ensures our model can generate fair and unbiased results.}.
Following~\citet{DBLP:conf/acl/WangHZ19}, we use the five public benchmark datasets in English for multi-way classification of lexical relations to evaluate KEML, namely, K\&H+N~\cite{DBLP:conf/starsem/NecsulescuMJBN15}, BLESS~\cite{baroni2011we}, ROOT09~\cite{DBLP:conf/lrec/SantusLCLH16}, EVALution~\cite{DBLP:conf/acl-ldl/SantusYLH15} and CogALex-V Subtask 2~\cite{DBLP:conf/cogalex/SantusGEL16}. Statistics of all the datasets are shown in Table~\ref{tab:stat}.
K\&H+N, BLESS, ROOT09 and EVALution are partitioned into training, validation and testing sets, following the exact same settings as in~\citet{DBLP:conf/cogalex/ShwartzD16}. The CogALex-V dataset has training and testing sets only, with no validation sets provided~\cite{DBLP:conf/cogalex/SantusGEL16}. Hence, we randomly sample 80\% of the training set to train the model, and use the rest for tuning. 

The default hyper-parameter settings of KEML are as follows:\footnote{We empirically set $N=\vert\mathcal{R}\vert-1$ to ensure that in each iteration of meta-learning, each auxiliary task is learned once in average.} $N=\vert\mathcal{R}\vert-1$, $\gamma=1$ and $\alpha=\epsilon=10^{-3}$. We use $tanh$ as the activation function, and Adam as the optimizer to train the neural network. All the model parameters are $l_2$-regularized, with the hyper-parameter $\lambda=10^{-3}$. The batch size is set as 256. The dimension of hidden layers is set as the same of $d$ (768 for the base BERT model). The number of parameters of the final neural classifier is around 7M to 24M, depending on the number of classes.  The algorithms are implemented with TensorFlow and trained with NVIDIA Tesla P100 GPU. The training time of LKB-BERT for all datasets is about 1.5 hours. Afterwards, only a few minutes are required to train the model for each dataset. For evaluation, we use Precision, Recall and F1 as metrics, reported as the average of all the classes, weighted by the support.

\subsection{General Experimental Results}

We first evaluate KEML over K\&H+N, BLESS, ROOT09 and EVALution. Since EVALution does not contain $\mathbf{RAN}$ relations, in the meta-learning process, we randomly sample relation triples from $\mathcal{D}$ that do not have the relation $r$, and take them as $\mathcal{S}_\mathbf{RAN}$. We manually tune the regularization hyper-parameter $\lambda$ from $10^{-2}$ to $10^{-4}$ using the validation set (based on F1) and report the performance over the testing set. As for baselines, we consider traditional distributional models Concat~\cite{DBLP:conf/eacl/BaroniBDS12} and Diff~\cite{DBLP:conf/coling/WeedsCRWK14}, pattern-based models NPB~\cite{DBLP:conf/acl/ShwartzGD16}, LexNET~\cite{DBLP:conf/cogalex/ShwartzD16},  NPB+Aug and LexNET+Aug~\cite{DBLP:conf/naacl/WashioK18}, and the state-of-the-art model SphereRE~\cite{DBLP:conf/acl/WangHZ19}. We refer readers to the following papers~\cite{DBLP:conf/cogalex/ShwartzD16,DBLP:conf/naacl/WashioK18,DBLP:conf/acl/WangHZ19} for the detailed descriptions of these baselines. Additionally, we implement two variants of our approach: i) LKB-BERT (using trained concept representations to predict lexical relations) and ii) KEML-S (KEML without the meta-learning stage). The results of KEML and all the baselines are summarized in Table~\ref{tab:f1}. 

As shown, KEML outperforms all baselines, especially over BLESS, ROOT09 and EVALution. As for K\&H+N, KEML produces a slightly better F1 score (0.3\%) than the strongest baseline SphereRE~\cite{DBLP:conf/acl/WangHZ19}. A probable cause is that K\&H+N is an ``easy'' dataset (99\% F1 by SphereRE), leaving little room for improvement.
Comparing KEML against LKB-BERT and KEML-S, we can conclude that, the knowledge enrichment technique for concept representation learning and meta-learning are highly beneficial for accurate prediction of lexical relations.

\subsection{Results of CogALex-V Shared Task}
		
We evaluate KEML over the CogALex-V Shared Task (Subtask 2)~\cite{DBLP:conf/cogalex/SantusGEL16}.
This dataset is most challenging as it contains a large number of random word pairs and disables ``lexical memorization''.
The organizer requires participants to discard the results of the random class from average, and report the F1 scores for each type of lexical relations.
We consider two top systems reported in this task (i.e., GHHH~\cite{DBLP:conf/cogalex/AttiaMSKS16} and LexNET~\cite{DBLP:conf/cogalex/ShwartzD16}), as well as two recent models that have been evaluated over the shared task (i.e., STM~\cite{DBLP:conf/naacl/GlavasV18} and SphereRE~\cite{DBLP:conf/acl/WangHZ19}) as strong competitors.
Because the training set contains an overwhelming number of random word pairs, during the training process of KEML-S and KEML, we randomly discard 70\% (manually tuned) of the random pairs in each epoch (which further  improves the performance of KEML-S and KEML by 1.8\% and 2.2\%, in terms of overall F1). 
Results are reported in Table~\ref{tab:cog}, showing that KEML achieves the highest F1 score of 50.0\%.
It also has highest scores on three types of lexical relations: synonymy (SYN), hypernymy (HYP) and meronymy (MER).


\begin{table}  
\centering
\begin{small} 
\begin{tabular}{lllllll}  
\hline 
\bf Method & SYN & ANT & HYP & MER & All\\
\hline
GHHH & 0.204 & 0.448 & 0.491 & 0.497 & 0.423\\
LexNET & 0.297 & 0.425 & 0.526 & 0.493 & 0.445\\
STM & 0.221 & \bf 0.504 & 0.498 & 0.504 & 0.453\\
SphereRE & 0.286 & 0.479 & 0.538 & 0.539 & 0.471\\
\hline 
\bf LKB-BERT & 0.281 & 0.470 & 0.532 & 0.530 & 0.464\\
\bf KEML-S & 0.276 & 0.470 & \bf 0.542 & \bf 0.631 & \bf 0.485\\
\bf KEML & \bf 0.292 & 0.492 & \bf 0.547 & \bf 0.652 & \bf 0.500\\
\hline
\end{tabular}
\end{small} 
\caption{LRC results for each lexical relation types over the CogALex-V shared task in terms of F1.} \label{tab:cog} 
\end{table}

\begin{figure*}
\centering
\subfigure[Dataset: K\&H+N]{
\includegraphics[width=0.25\textwidth]{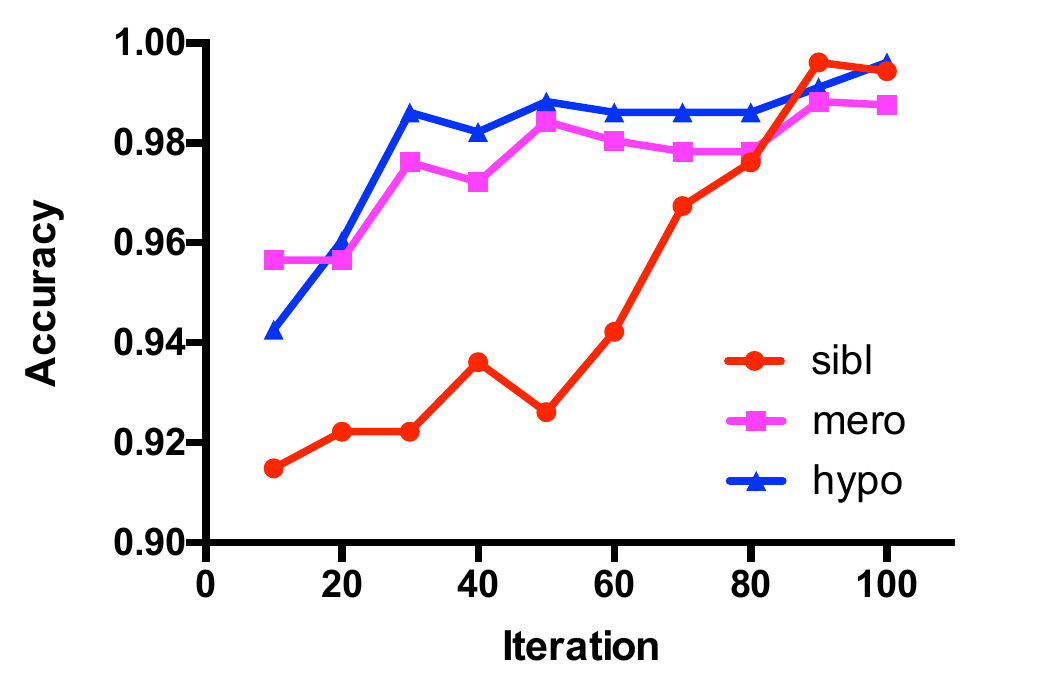}}
\subfigure[Dataset: BLESS]{
\includegraphics[width=0.25\textwidth]{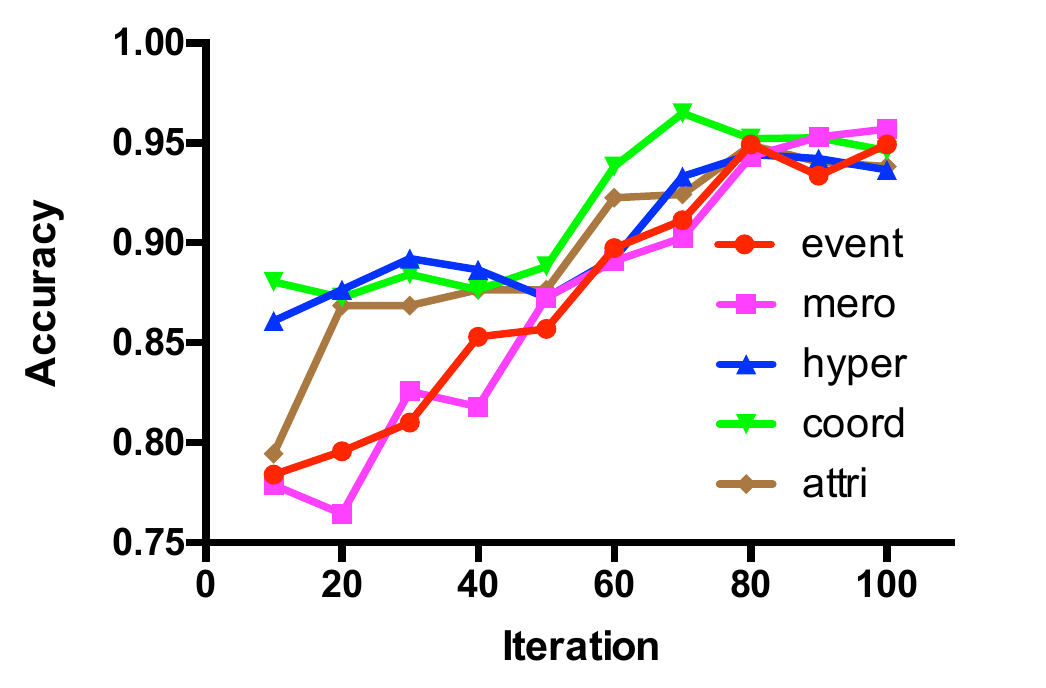}}
\subfigure[Dataset: ROOT09]{
\includegraphics[width=0.25\textwidth]{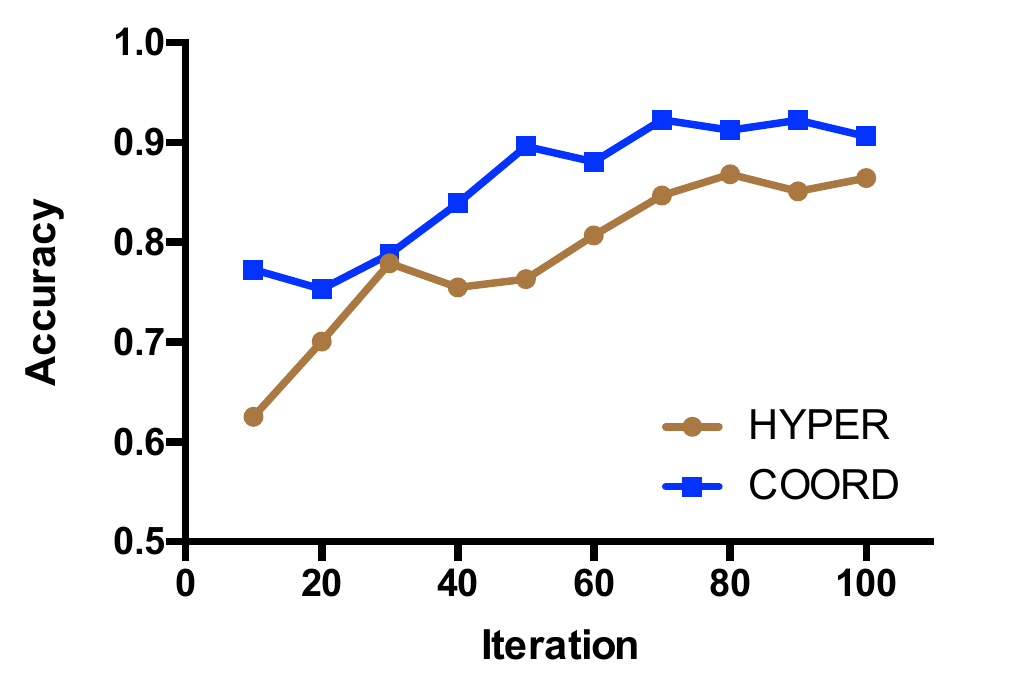}}
\caption{Accuracy of single relation prediction during the meta-learning process (best viewed in color).} \label{tab:acc} 
\label{fig:meta}
\end{figure*}

\subsection{Detailed Analysis of KEML}
To facilitate deeper understanding, we conduct additional experiments to analyze KEML's components.
We first study how knowledge-enriched concept representation learning benefits LRC. We implement three models: LKB-BERT (Binary), LKB-BERT (Multi) and LKB-BERT (Full). LKB-BERT (Binary) and LKB-BERT (Multi) only fine-tune on single objective: $\mathcal{L}_{KB}^{(2)}$ and $\mathcal{L}_{KB}^{(1)}$, respectively. LKB-BERT (Full) is the full implementation, as described previously. We take the two concepts' representations as features ($\vec{x}_i$ and $\vec{y}_i$) to train relation classifiers for LRC and report the results over the validation sets. For fair comparison, we use neural networks with one hidden layer (with the dimension $d$, and the activation function $tanh$)  as classifiers in all the experiments and present the results in Table~\ref{tab:bert}. 
We can see that the improvement of using $\mathcal{L}_{KB}=\mathcal{L}_{KB}^{(2)}+\mathcal{L}_{KB}^{(1)}$ as the objective function is consistent across all the datasets (which ranges  from 0.8\% to 1.9\% in terms of F1).
Particularly, LKB-BERT outperforms the strongest baseline SphereRE in a few cases (for example, the dataset ROOT09, as shown in Table~\ref{tab:f1} and Table~\ref{tab:bert}). Hence, LKB-BERT is capable of encoding knowledge in patterns and lexical knowledge bases to represent the semantics of lexical relations.

\begin{table}  
\begin{small} 
\begin{center}
\begin{tabular}{lllllll}  
\hline 
\bf Dataset & \bf Binary & \bf Multi & \bf Full\\
\hline
K\&H+N & 0.964 & 0.972 & \bf 0.983\\
BLESS & 0.921 & 0.929 & \bf 0.939\\
ROOT09 & 0.854 & 0.861 & \bf 0.863\\
EVALution & 0.630 & 0.632 & \bf 0.641\\
CogALex-V & 0.464 & 0.467 & \bf 0.472\\
\hline
\end{tabular}
\end{center}
\end{small} 
\caption{LRC results using concept embeddings generated by LKB-BERT and variants in terms of F1.} \label{tab:bert} 
\end{table}

Next, we look into the meta-learning process in KEML. We test whether SRR Cells can distinguish a specific type of lexical relations from random concept pairs. In each iteration of meta-learning, we sample another batch of positive and negative samples from $\mathcal{D}$ and compute the accuracy of single relation recognition. Figure~\ref{fig:meta} illustrates how accuracy changes through time in K\&H+N, BLESS and ROOT09. Within 100 iterations, our models can achieve desirable performance efficiently, achieving good parameter initializations for LRC. This experiment also explains why KEML produces better results than KEML-S.

\subsection{Error Analysis}

We analyze prediction errors produced by KEML. Because the inputs of our task are very simple and the interpretation of deep neural language models is still challenging, the error analysis process is rather difficult. Here, we analyze the causes of errors from a linguistic point of view, with some cases presented in Table~\ref{tab:acc}. As seen, some types of lexical relations are very similar in semantics. For instance, concept pairs with the \emph{synonymy} relation and the \emph{co-hyponymy} relation are usually mapped similar positions in the embedding space. Hence, it is difficult for models to distinguish the differences between the two relations without rich contextual information available. Another problem is that some of the relations are ``blurry'' in semantics, making KEML hard to discriminate between these relations and random relations. 

\begin{table}  
\begin{small} 
\begin{center}
\begin{tabular}{lll}  
\hline 
\bf Concept Pairs & \bf Predicted & \bf True\\
\hline
(turtle, frog) & Synonym & Co-hyponym \\
(draw, pull) & Random & Synonym\\
(bowl, glass) & Co-hyponymy & Meronymy\\
(cannon, warrior) & Synonym & Random\\
(affection, healthy) & Co-hyponym & Attribute\\
\hline  
\end{tabular}
\end{center}
\end{small} 
\caption{Cases of prediction errors.} \label{tab:error} 
\end{table}  

\begin{figure}
\centering
\includegraphics[width=0.4\textwidth]{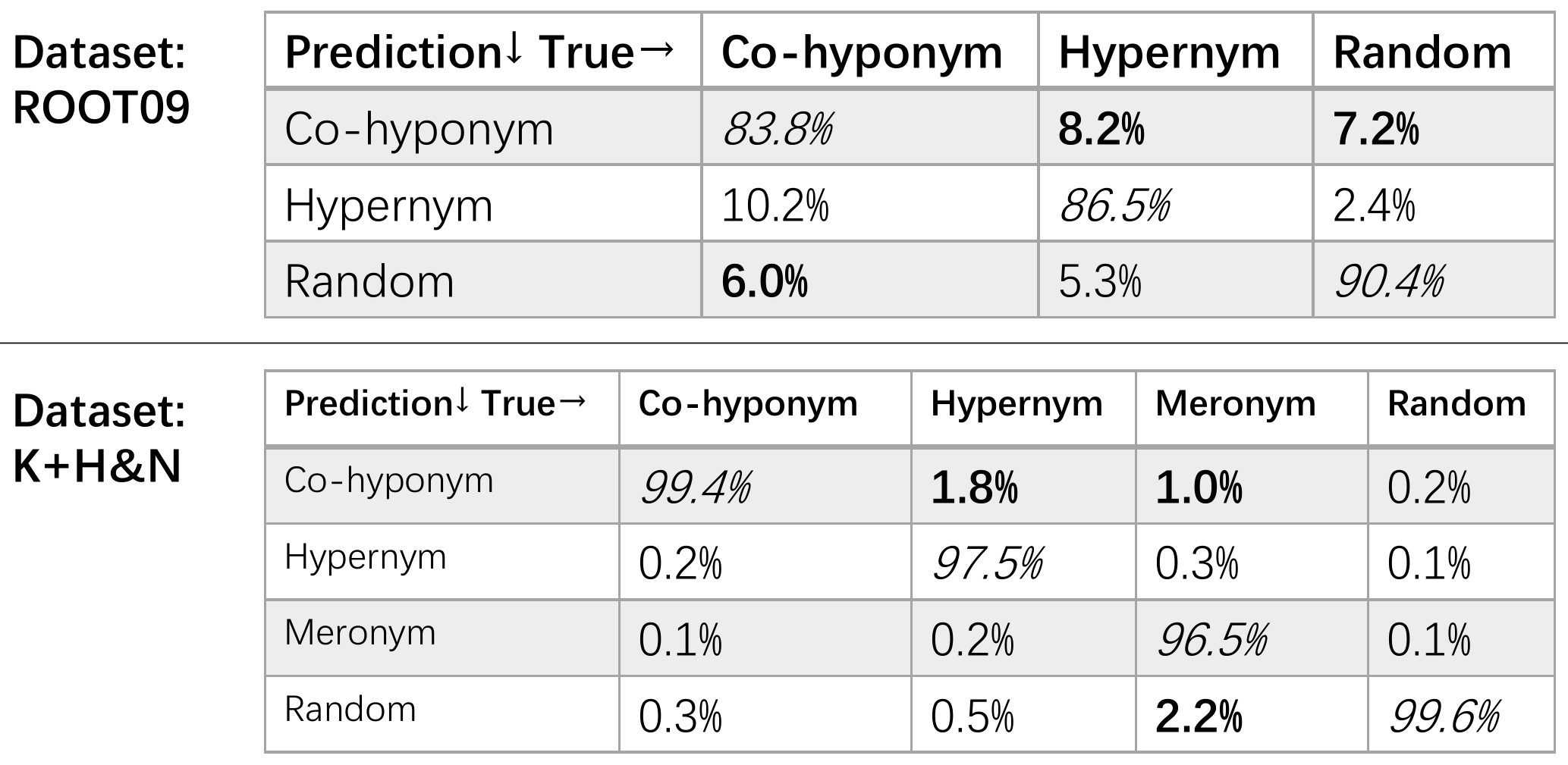}
\caption{Inter-class prediction error analysis on ROOT09 and K+H\&N. Top-3 error categories are marked in bold.} 
\label{fig:matrix}
\end{figure}

Additionally, we analyze the percentages of testing instances of each class being mis-classified to other classes. The results over ROOT09 and K+H\&N are shown in Figure~\ref{fig:matrix}. Take K+H\&N as an example. Although the prediction results are already highly accurate, for classes with few training instances (e.g., meronym), there is still room for further improvement, which will be left as future work.

\section{Conclusion and Future Work}
\label{sec:con}

In this paper, we present the Knowledge-Enriched Meta-Learning (KEML) framework to distinguish different lexical relations.
Experimental results confirm that KEML outperforms state-of-the-art approaches.
Future work includes: i) improving relation representation learning with deep neural language models~\cite{DBLP:conf/acl/VylomovaRCB16,DBLP:conf/acl/AnkeSW19,DBLP:conf/ijcai/Camacho-Collados19}; ii) integrating richer linguistic and commonsense knowledge into KEML; and iii) applying KEML to downstream tasks such as taxonomy learning~\cite{DBLP:conf/semeval/BordeaLB16,DBLP:conf/semeval/JurgensP16,DBLP:journals/kais/WangFHZ19}.


\begin{thebibliography}{62}
\providecommand{\natexlab}[1]{#1}
\providecommand{\url}[1]{\texttt{#1}}
\providecommand{\urlprefix}{URL }
\expandafter\ifx\csname urlstyle\endcsname\relax
  \providecommand{\doi}[1]{doi:\discretionary{}{}{}#1}\else
  \providecommand{\doi}{doi:\discretionary{}{}{}\begingroup
  \urlstyle{rm}\Url}\fi

\bibitem[{Anke, Schockaert, and Wanner(2019)}]{DBLP:conf/acl/AnkeSW19}
Anke, L.~E.; Schockaert, S.; and Wanner, L. 2019.
\newblock Collocation Classification with Unsupervised Relation Vectors.
\newblock In \emph{ACL}, 5765--5772.

\bibitem[{Attia et~al.(2016)Attia, Maharjan, Samih, Kallmeyer, and
  Solorio}]{DBLP:conf/cogalex/AttiaMSKS16}
Attia, M.; Maharjan, S.; Samih, Y.; Kallmeyer, L.; and Solorio, T. 2016.
\newblock CogALex-V Shared Task: {GHHH} - Detecting Semantic Relations via Word
  Embeddings.
\newblock In \emph{CogALex@COLING}, 86--91.

\bibitem[{Baroni et~al.(2012)Baroni, Bernardi, Do, and
  Shan}]{DBLP:conf/eacl/BaroniBDS12}
Baroni, M.; Bernardi, R.; Do, N.; and Shan, C. 2012.
\newblock Entailment above the word level in distributional semantics.
\newblock In \emph{EACL}, 23--32.

\bibitem[{Baroni and Lenci(2011)}]{baroni2011we}
Baroni, M.; and Lenci, A. 2011.
\newblock How we BLESSed distributional semantic evaluation.
\newblock In \emph{GEMS Workshop}.

\bibitem[{Bordea, Lefever, and Buitelaar(2016)}]{DBLP:conf/semeval/BordeaLB16}
Bordea, G.; Lefever, E.; and Buitelaar, P. 2016.
\newblock SemEval-2016 Task 13: Taxonomy Extraction Evaluation (TExEval-2).
\newblock In \emph{SemEval@NAACL-HLT}, 1081--1091.

\bibitem[{Camacho{-}Collados et~al.(2019)Camacho{-}Collados, Anke, Jameel, and
  Schockaert}]{DBLP:conf/ijcai/Camacho-Collados19}
Camacho{-}Collados, J.; Anke, L.~E.; Jameel, S.; and Schockaert, S. 2019.
\newblock A Latent Variable Model for Learning Distributional Relation Vectors.
\newblock In Kraus, S., ed., \emph{IJCAI}, 4911--4917.

\bibitem[{Camacho{-}Collados, Anke, and
  Schockaert(2019)}]{DBLP:conf/acl/Camacho-Collados19}
Camacho{-}Collados, J.; Anke, L.~E.; and Schockaert, S. 2019.
\newblock Relational Word Embeddings.
\newblock In \emph{ACL}, 3286--3296.

\bibitem[{Chen et~al.(2019)Chen, Zhang, Zhang, Chen, and
  Chen}]{DBLP:conf/emnlp/ChenZZCC19}
Chen, M.; Zhang, W.; Zhang, W.; Chen, Q.; and Chen, H. 2019.
\newblock Meta Relational Learning for Few-Shot Link Prediction in Knowledge
  Graphs.
\newblock In \emph{EMNLP-IJCNLP}, 4216--4225.

\bibitem[{Cordeiro et~al.(2016)Cordeiro, Ramisch, Idiart, and
  Villavicencio}]{DBLP:conf/acl/CordeiroRIV16}
Cordeiro, S.; Ramisch, C.; Idiart, M.; and Villavicencio, A. 2016.
\newblock Predicting the Compositionality of Nominal Compounds: Giving Word
  Embeddings a Hard Time.
\newblock In \emph{ACL}, 1986–1997.

\bibitem[{Dai et~al.(2019)Dai, Yang, Yang, Carbonell, Le, and
  Salakhutdinov}]{DBLP:conf/acl/DaiYYCLS19}
Dai, Z.; Yang, Z.; Yang, Y.; Carbonell, J.~G.; Le, Q.~V.; and Salakhutdinov, R.
  2019.
\newblock Transformer-XL: Attentive Language Models beyond a Fixed-Length
  Context.
\newblock In \emph{ACL}, 2978--2988.

\bibitem[{Devlin et~al.(2019)Devlin, Chang, Lee, and
  Toutanova}]{DBLP:conf/naacl/DevlinCLT19}
Devlin, J.; Chang, M.; Lee, K.; and Toutanova, K. 2019.
\newblock {BERT:} Pre-training of Deep Bidirectional Transformers for Language
  Understanding.
\newblock In \emph{NAACL}, 4171--4186.

\bibitem[{Dou, Yu, and Anastasopoulos(2019)}]{DBLP:conf/emnlp/DouYA19}
Dou, Z.; Yu, K.; and Anastasopoulos, A. 2019.
\newblock Investigating Meta-Learning Algorithms for Low-Resource Natural
  Language Understanding Tasks.
\newblock In \emph{EMNLP-IJCNLP}, 1192--1197.

\bibitem[{Finn, Abbeel, and Levine(2017)}]{DBLP:conf/icml/FinnAL17}
Finn, C.; Abbeel, P.; and Levine, S. 2017.
\newblock Model-Agnostic Meta-Learning for Fast Adaptation of Deep Networks.
\newblock In \emph{ICML}, volume~70, 1126--1135.

\bibitem[{Finn, Xu, and Levine(2018)}]{DBLP:conf/nips/FinnXL18}
Finn, C.; Xu, K.; and Levine, S. 2018.
\newblock Probabilistic Model-Agnostic Meta-Learning.
\newblock In \emph{NeurIPS}, 9537--9548.

\bibitem[{Fu et~al.(2014)Fu, Guo, Qin, Che, Wang, and
  Liu}]{DBLP:conf/acl/FuGQCWL14}
Fu, R.; Guo, J.; Qin, B.; Che, W.; Wang, H.; and Liu, T. 2014.
\newblock Learning Semantic Hierarchies via Word Embeddings.
\newblock In \emph{ACL}, 1199--1209.

\bibitem[{Glavas and Vulic(2018)}]{DBLP:conf/naacl/GlavasV18}
Glavas, G.; and Vulic, I. 2018.
\newblock Discriminating between Lexico-Semantic Relations with the
  Specialization Tensor Model.
\newblock In \emph{NAACL}, 181--187.

\bibitem[{Guo et~al.(2019)Guo, Tang, Duan, Zhou, and
  Yin}]{DBLP:conf/acl/GuoTDZY19}
Guo, D.; Tang, D.; Duan, N.; Zhou, M.; and Yin, J. 2019.
\newblock Coupling Retrieval and Meta-Learning for Context-Dependent Semantic
  Parsing.
\newblock In \emph{ACL}, 855--866.

\bibitem[{He et~al.(2016)He, Zhang, Ren, and Sun}]{DBLP:conf/cvpr/HeZRS16}
He, K.; Zhang, X.; Ren, S.; and Sun, J. 2016.
\newblock Deep Residual Learning for Image Recognition.
\newblock In \emph{CVPR}, 770--778.

\bibitem[{Hearst(1992)}]{DBLP:conf/coling/Hearst92}
Hearst, M.~A. 1992.
\newblock Automatic Acquisition of Hyponyms from Large Text Corpora.
\newblock In \emph{COLING}, 539--545.

\bibitem[{Joshi et~al.(2019)Joshi, Choi, Levy, Weld, and
  Zettlemoyer}]{DBLP:conf/naacl/JoshiCLWZ19}
Joshi, M.; Choi, E.; Levy, O.; Weld, D.~S.; and Zettlemoyer, L. 2019.
\newblock pair2vec: Compositional Word-Pair Embeddings for Cross-Sentence
  Inference.
\newblock In \emph{NAACL}, 3597--3608.

\bibitem[{Jurgens and Pilehvar(2016)}]{DBLP:conf/semeval/JurgensP16}
Jurgens, D.; and Pilehvar, M.~T. 2016.
\newblock SemEval-2016 Task 14: Semantic Taxonomy Enrichment.
\newblock In \emph{SemEval@NAACL-HLT}, 1092--1102.

\bibitem[{Lan et~al.(2019)Lan, Chen, Goodman, Gimpel, Sharma, and
  Soricut}]{DBLP:journals/corr/abs-1909-11942}
Lan, Z.; Chen, M.; Goodman, S.; Gimpel, K.; Sharma, P.; and Soricut, R. 2019.
\newblock {ALBERT:} {A} Lite {BERT} for Self-supervised Learning of Language
  Representations.
\newblock \emph{CoRR} abs/1909.11942.

\bibitem[{Le et~al.(2019)Le, Roller, Papaxanthos, Kiela, and
  Nickel}]{DBLP:conf/acl/LeRPKN19}
Le, M.; Roller, S.; Papaxanthos, L.; Kiela, D.; and Nickel, M. 2019.
\newblock Inferring Concept Hierarchies from Text Corpora via Hyperbolic
  Embeddings.
\newblock In \emph{ACL}, 3231--3241.

\bibitem[{Levy et~al.(2015)Levy, Remus, Biemann, and
  Dagan}]{DBLP:conf/naacl/LevyRBD15}
Levy, O.; Remus, S.; Biemann, C.; and Dagan, I. 2015.
\newblock Do Supervised Distributional Methods Really Learn Lexical Inference
  Relations?
\newblock In \emph{NAACL}, 970--976.

\bibitem[{Liu et~al.(2017)Liu, Ren, Zhu, Zhi, Gui, Ji, and
  Han}]{DBLP:conf/emnlp/LiuRZZGJH17}
Liu, L.; Ren, X.; Zhu, Q.; Zhi, S.; Gui, H.; Ji, H.; and Han, J. 2017.
\newblock Heterogeneous Supervision for Relation Extraction: {A} Representation
  Learning Approach.
\newblock In \emph{EMNLP}, 46--56.

\bibitem[{Liu et~al.(2019)Liu, Zhou, Zhao, Wang, Ju, Deng, and
  Wang}]{DBLP:journals/corr/abs-1909-07606}
Liu, W.; Zhou, P.; Zhao, Z.; Wang, Z.; Ju, Q.; Deng, H.; and Wang, P. 2019.
\newblock {K-BERT:} Enabling Language Representation with Knowledge Graph.
\newblock \emph{CoRR} abs/1909.07606.

\bibitem[{Madotto et~al.(2019)Madotto, Lin, Wu, and
  Fung}]{DBLP:conf/acl/MadottoLWF19}
Madotto, A.; Lin, Z.; Wu, C.; and Fung, P. 2019.
\newblock Personalizing Dialogue Agents via Meta-Learning.
\newblock In \emph{ACL}, 5454--5459.

\bibitem[{Mikolov et~al.(2013)Mikolov, Chen, Corrado, and
  Dean}]{DBLP:journals/corr/abs-1301-3781}
Mikolov, T.; Chen, K.; Corrado, G.; and Dean, J. 2013.
\newblock Efficient Estimation of Word Representations in Vector Space.
\newblock In \emph{ICLR}.

\bibitem[{Miller(1995)}]{DBLP:journals/cacm/Miller95}
Miller, G.~A. 1995.
\newblock WordNet: {A} Lexical Database for English.
\newblock \emph{Commun. {ACM}} 38(11): 39--41.
\newblock \doi{10.1145/219717.219748}.

\bibitem[{Mrksic et~al.(2017)Mrksic, Vulic, S{\'{e}}aghdha, Leviant, Reichart,
  Gasic, Korhonen, and Young}]{DBLP:journals/tacl/MrksicVSLRGKY17}
Mrksic, N.; Vulic, I.; S{\'{e}}aghdha, D.~{\'{O}}.; Leviant, I.; Reichart, R.;
  Gasic, M.; Korhonen, A.; and Young, S.~J. 2017.
\newblock Semantic Specialization of Distributional Word Vector Spaces using
  Monolingual and Cross-Lingual Constraints.
\newblock \emph{{TACL}} 5: 309--324.

\bibitem[{Necsulescu et~al.(2015)Necsulescu, Mendes, Jurgens, Bel, and
  Navigli}]{DBLP:conf/starsem/NecsulescuMJBN15}
Necsulescu, S.; Mendes, S.; Jurgens, D.; Bel, N.; and Navigli, R. 2015.
\newblock Reading Between the Lines: Overcoming Data Sparsity for Accurate
  Classification of Lexical Relationships.
\newblock In \emph{*SEM}, 182--192.

\bibitem[{Nguyen, {Schulte im Walde}, and Vu(2016)}]{DBLP:conf/acl/NguyenWV16}
Nguyen, K.~A.; {Schulte im Walde}, S.; and Vu, N.~T. 2016.
\newblock Integrating Distributional Lexical Contrast into Word Embeddings for
  Antonym-Synonym Distinction.
\newblock In \emph{ACL}.

\bibitem[{Nguyen, {Schulte im Walde}, and Vu(2017)}]{DBLP:conf/eacl/NguyenWV17}
Nguyen, K.~A.; {Schulte im Walde}, S.; and Vu, N.~T. 2017.
\newblock Distinguishing Antonyms and Synonyms in a Pattern-based Neural
  Network.
\newblock In \emph{EACL}, 76--85.

\bibitem[{Nichol, Achiam, and
  Schulman(2018)}]{DBLP:journals/corr/abs-1803-02999}
Nichol, A.; Achiam, J.; and Schulman, J. 2018.
\newblock On First-Order Meta-Learning Algorithms.
\newblock \emph{CoRR} abs/1803.02999.

\bibitem[{Peters et~al.(2018)Peters, Neumann, Iyyer, Gardner, Clark, Lee, and
  Zettlemoyer}]{DBLP:conf/naacl/PetersNIGCLZ18}
Peters, M.~E.; Neumann, M.; Iyyer, M.; Gardner, M.; Clark, C.; Lee, K.; and
  Zettlemoyer, L. 2018.
\newblock Deep Contextualized Word Representations.
\newblock In \emph{NAACL}, 2227--2237.

\bibitem[{Ponti et~al.(2019)Ponti, Vulic, Glavas, Reichart, and
  Korhonen}]{DBLP:conf/emnlp/PontiVGRK19}
Ponti, E.~M.; Vulic, I.; Glavas, G.; Reichart, R.; and Korhonen, A. 2019.
\newblock Cross-lingual Semantic Specialization via Lexical Relation Induction.
\newblock In \emph{EMNLP-IJCNLP}, 2206--2217.

\bibitem[{Qiu et~al.(2020)Qiu, Sun, Xu, Shao, Dai, and
  Huang}]{DBLP:journals/corr/abs-2003-08271}
Qiu, X.; Sun, T.; Xu, Y.; Shao, Y.; Dai, N.; and Huang, X. 2020.
\newblock Pre-trained Models for Natural Language Processing: {A} Survey.
\newblock \emph{CoRR} abs/2003.08271.
\newblock \urlprefix\url{https://arxiv.org/abs/2003.08271}.

\bibitem[{Roller and Erk(2016)}]{DBLP:conf/emnlp/RollerE16}
Roller, S.; and Erk, K. 2016.
\newblock Relations such as Hypernymy: Identifying and Exploiting Hearst
  Patterns in Distributional Vectors for Lexical Entailment.
\newblock In \emph{EMNLP}, 2163--2172.

\bibitem[{Roller, Kiela, and Nickel(2018)}]{DBLP:conf/acl/RollerKN18}
Roller, S.; Kiela, D.; and Nickel, M. 2018.
\newblock Hearst Patterns Revisited: Automatic Hypernym Detection from Large
  Text Corpora.
\newblock In \emph{ACL}, 358--363.

\bibitem[{Santus et~al.(2016{\natexlab{a}})Santus, Gladkova, Evert, and
  Lenci}]{DBLP:conf/cogalex/SantusGEL16}
Santus, E.; Gladkova, A.; Evert, S.; and Lenci, A. 2016{\natexlab{a}}.
\newblock The CogALex-V Shared Task on the Corpus-Based Identification of
  Semantic Relations.
\newblock In \emph{CogALex@COLING}, 69--79.

\bibitem[{Santus et~al.(2016{\natexlab{b}})Santus, Lenci, Chiu, Lu, and
  Huang}]{DBLP:conf/lrec/SantusLCLH16}
Santus, E.; Lenci, A.; Chiu, T.; Lu, Q.; and Huang, C. 2016{\natexlab{b}}.
\newblock Nine Features in a Random Forest to Learn Taxonomical Semantic
  Relations.
\newblock In \emph{LREC}.

\bibitem[{Santus et~al.(2015)Santus, Yung, Lenci, and
  Huang}]{DBLP:conf/acl-ldl/SantusYLH15}
Santus, E.; Yung, F.; Lenci, A.; and Huang, C. 2015.
\newblock EVALution 1.0: an Evolving Semantic Dataset for Training and
  Evaluation of Distributional Semantic Models.
\newblock In \emph{LDL@ACL-IJCNLP}, 64--69.

\bibitem[{Shen et~al.(2018)Shen, Wu, Lei, Zhang, Ren, Vanni, Sadler, and
  Han}]{DBLP:conf/kdd/ShenWLZRVS018}
Shen, J.; Wu, Z.; Lei, D.; Zhang, C.; Ren, X.; Vanni, M.~T.; Sadler, B.~M.; and
  Han, J. 2018.
\newblock HiExpan: Task-Guided Taxonomy Construction by Hierarchical Tree
  Expansion.
\newblock In \emph{KDD}, 2180--2189.

\bibitem[{Shwartz and Dagan(2016)}]{DBLP:conf/cogalex/ShwartzD16}
Shwartz, V.; and Dagan, I. 2016.
\newblock Path-based vs. Distributional Information in Recognizing Lexical
  Semantic Relations.
\newblock In \emph{CogALex@COLING}, 24--29.

\bibitem[{Shwartz, Goldberg, and Dagan(2016)}]{DBLP:conf/acl/ShwartzGD16}
Shwartz, V.; Goldberg, Y.; and Dagan, I. 2016.
\newblock Improving Hypernymy Detection with an Integrated Path-based and
  Distributional Method.
\newblock In \emph{ACL}.

\bibitem[{Thompson et~al.(2019)Thompson, Knowles, Zhang, Khayrallah, Duh, and
  Koehn}]{DBLP:conf/emnlp/ThompsonKZKDK19}
Thompson, B.; Knowles, R.; Zhang, X.; Khayrallah, H.; Duh, K.; and Koehn, P.
  2019.
\newblock HABLex: Human Annotated Bilingual Lexicons for Experiments in Machine
  Translation.
\newblock In \emph{EMNLP-IJCNLP}, 1382--1387.

\bibitem[{Vanschoren(2018)}]{DBLP:journals/corr/abs-1810-03548}
Vanschoren, J. 2018.
\newblock Meta-Learning: {A} Survey.
\newblock \emph{CoRR} abs/1810.03548.

\bibitem[{Vylomova et~al.(2016)Vylomova, Rimell, Cohn, and
  Baldwin}]{DBLP:conf/acl/VylomovaRCB16}
Vylomova, E.; Rimell, L.; Cohn, T.; and Baldwin, T. 2016.
\newblock Take and Took, Gaggle and Goose, Book and Read: Evaluating the
  Utility of Vector Differences for Lexical Relation Learning.
\newblock In \emph{ACL}, 1671--1682.

\bibitem[{Wang et~al.(2019)Wang, Fan, He, and
  Zhou}]{DBLP:journals/kais/WangFHZ19}
Wang, C.; Fan, Y.; He, X.; and Zhou, A. 2019.
\newblock Predicting hypernym-hyponym relations for Chinese taxonomy learning.
\newblock \emph{Knowl. Inf. Syst.} 58(3): 585--610.

\bibitem[{Wang, He, and Zhou(2017)}]{DBLP:conf/emnlp/WangHZ17}
Wang, C.; He, X.; and Zhou, A. 2017.
\newblock A Short Survey on Taxonomy Learning from Text Corpora: Issues,
  Resources and Recent Advances.
\newblock In \emph{EMNLP}, 1190--1203.

\bibitem[{Wang, He, and Zhou(2019)}]{DBLP:conf/acl/WangHZ19}
Wang, C.; He, X.; and Zhou, A. 2019.
\newblock SphereRE: Distinguishing Lexical Relations with Hyperspherical
  Relation Embeddings.
\newblock In \emph{ACL}, 1727--1737.

\bibitem[{Wang et~al.(2017)Wang, Mao, Wang, and
  Guo}]{DBLP:journals/tkde/WangMWG17}
Wang, Q.; Mao, Z.; Wang, B.; and Guo, L. 2017.
\newblock Knowledge Graph Embedding: {A} Survey of Approaches and Applications.
\newblock \emph{{IEEE} Trans. Knowl. Data Eng.} 29(12): 2724--2743.
\newblock \doi{10.1109/TKDE.2017.2754499}.

\bibitem[{Washio and Kato(2018{\natexlab{a}})}]{DBLP:conf/naacl/WashioK18}
Washio, K.; and Kato, T. 2018{\natexlab{a}}.
\newblock Filling Missing Paths: Modeling Co-occurrences of Word Pairs and
  Dependency Paths for Recognizing Lexical Semantic Relations.
\newblock In \emph{NAACL}, 1123--1133.

\bibitem[{Washio and Kato(2018{\natexlab{b}})}]{DBLP:conf/emnlp/WashioK18}
Washio, K.; and Kato, T. 2018{\natexlab{b}}.
\newblock Neural Latent Relational Analysis to Capture Lexical Semantic
  Relations in a Vector Space.
\newblock In \emph{EMNLP}, 594--600.

\bibitem[{Weeds et~al.(2014)Weeds, Clarke, Reffin, Weir, and
  Keller}]{DBLP:conf/coling/WeedsCRWK14}
Weeds, J.; Clarke, D.; Reffin, J.; Weir, D.~J.; and Keller, B. 2014.
\newblock Learning to Distinguish Hypernyms and Co-Hyponyms.
\newblock In \emph{COLING}, 2249--2259.

\bibitem[{Wu and He(2019)}]{DBLP:conf/cikm/WuH19a}
Wu, S.; and He, Y. 2019.
\newblock Enriching Pre-trained Language Model with Entity Information for
  Relation Classification.
\newblock In \emph{CIKM}, 2361--2364.

\bibitem[{Wu et~al.(2012)Wu, Li, Wang, and Zhu}]{DBLP:conf/sigmod/WuLWZ12}
Wu, W.; Li, H.; Wang, H.; and Zhu, K.~Q. 2012.
\newblock Probase: a probabilistic taxonomy for text understanding.
\newblock In \emph{SIGMOD}, 481--492.

\bibitem[{Xiong et~al.(2020)Xiong, Du, Wang, and
  Stoyanov}]{DBLP:conf/iclr/XiongDWS20}
Xiong, W.; Du, J.; Wang, W.~Y.; and Stoyanov, V. 2020.
\newblock Pretrained Encyclopedia: Weakly Supervised Knowledge-Pretrained
  Language Model.
\newblock In \emph{ICLR}.

\bibitem[{Yang et~al.(2017)Yang, Zou, Wang, Yan, and
  Wen}]{DBLP:conf/aaai/YangZWYW17}
Yang, S.; Zou, L.; Wang, Z.; Yan, J.; and Wen, J. 2017.
\newblock Efficiently Answering Technical Questions - {A} Knowledge Graph
  Approach.
\newblock In \emph{AAAI}, 3111--3118.

\bibitem[{Yang et~al.(2019)Yang, Dai, Yang, Carbonell, Salakhutdinov, and
  Le}]{DBLP:conf/nips/YangDYCSL19}
Yang, Z.; Dai, Z.; Yang, Y.; Carbonell, J.~G.; Salakhutdinov, R.; and Le, Q.~V.
  2019.
\newblock XLNet: Generalized Autoregressive Pretraining for Language
  Understanding.
\newblock In \emph{NeurIPS}, 5754--5764.

\bibitem[{Yao, Mao, and Luo(2019)}]{DBLP:journals/corr/abs-1909-03193}
Yao, L.; Mao, C.; and Luo, Y. 2019.
\newblock {KG-BERT:} {BERT} for Knowledge Graph Completion.
\newblock \emph{CoRR} abs/1909.03193.

\bibitem[{Zhang et~al.(2019)Zhang, Han, Liu, Jiang, Sun, and
  Liu}]{DBLP:conf/acl/ZhangHLJSL19}
Zhang, Z.; Han, X.; Liu, Z.; Jiang, X.; Sun, M.; and Liu, Q. 2019.
\newblock {ERNIE:} Enhanced Language Representation with Informative Entities.
\newblock In \emph{ACL}, 1441--1451.

\end{thebibliography}
\end{document}